\documentclass{article}
\usepackage[margin=1in]{geometry}

\usepackage[separate-uncertainty = true,multi-part-units=single]{siunitx}
\usepackage{amsmath, array, url}
\usepackage{booktabs, multirow}

\usepackage{tikz}
\usetikzlibrary{external, automata, positioning}
\usetikzlibrary{shapes.geometric, arrows}
\tikzstyle{block2} = [rectangle, rounded corners, text centered, draw=black, text width=1.5cm, minimum height=0.75cm, line width=1pt, inner ysep=.25em]
\tikzstyle{arrow} = [thick,->,>=stealth]

\usepackage{subcaption}
\title{Speech-Based Human-Exoskeleton Interaction for\\ Lower Limb Motion Planning}

\author{Eddie Guo$^{1,2}$, Christopher Perlette$^3$, Mojtaba Sharifi$^{1,4,5}$, Lukas Grasse$^3$, Matthew Tata$^3$,\\ Vivian K. Mushahwar$^5$, and Mahdi Tavakoli$^{1,4,5}$\\[1em] \small $^1$Department of Electrical and Computer Engineering, University of Alberta,\\ \small $^2$Cumming School of Medicine, University of Calgary\\ \small $^3$Department of Neuroscience, University of Lethbridge\\ \small $^4$Department of Mechanical Engineering, San Jose State University\\ \small $^5$Sensory Motor Adaptive Rehabilitation Technology (SMART) Network, University of Alberta}

\begin{document}

\maketitle

\begin{abstract}
This study presents a speech-based motion planning strategy (SBMP) developed for lower limb exoskeletons to facilitate safe and compliant human-robot interaction. A speech processing system, finite state machine, and central pattern generator are the building blocks of the proposed strategy for online planning of the exoskeleton’s trajectory. According to experimental evaluations, this speech-processing system achieved low levels of word and intent errors. Regarding locomotion, the completion time for users with voice commands was 54\% faster than that using a mobile app interface. With the proposed SBMP, users are able to maintain their postural stability with both hands-free. This supports its use as an effective motion planning method for the assistance and rehabilitation of individuals with lower-limb impairments. \\

\end{abstract}

\section{Introduction}
Assistive robotic systems can enhance the quality of life of people affected by neurological impairments \cite{neuro-impair}. These systems include lower limb exoskeletons, such as ReWalk \cite{rewalk}, Indego \cite{indego}, HAL \cite{hal}, and Exo-H3 (used in these experiments) \cite{exo-h3}, which are designed to rehabilitate individuals with neurological impairments. In comparison with traditional physical therapies, wearable exoskeletons allow users to interact more easily with their environment, improving mobility and independence in non-ambulatory individuals \cite{neuro-impair}.

To facilitate safe and compliant human-robot interactions (HRIs), the exoskeleton motion planning strategy should be intuitive and efficient to use \cite{VILLANI2018}. Social HRI, where humans use body language, gestures, and speech to interact with robots, shows promise in addressing issues with physical interactions through quick and efficient identification of user intentions \cite{human-robotInteractionStatus}.

Although speech recognition (SR) is increasingly adopted for HRI, this adoption is generally limited to humanoid robots \cite{human-robotInteractionStatus, humanHRI}. Nonetheless, the integration of SR into other forms of robotics may improve their ergonomics and practicality for human use. For example, SR reduces the need for conventional interaction methods such as button-based interfaces or mobile apps, allowing for hands-free use of an exoskeleton. This aspect of SR benefits users who require their hands to grasp stability aids, such as crutches or a walker. Furthermore, it increases safety in emergencies, where users may find voice commands more intuitive than a tactile controller. However, these interactions are only effective if the robot can perceive what a user is saying.

Computational SR is a complex task that requires turning auditory information into text. Over the last 40 years, there have been several breakthroughs that have advanced the technology to the point where it can effectively analyze speech and produce an accurate response even under adverse conditions \cite{speech-recognition-status}. In particular, a neural network architecture known as the time delay neural network factorization model has been shown to be effective at acoustic modeling and speech perception \cite{TDNNFEfficacy, moritz2016acoustic}.

Natural language understanding (NLU) addresses the problem of interpreting user intentions through two primary methods. The first method (true NLU) uses a combination of keyword analysis, semantic processing, discourse processing, and context analysis to determine the meaning and intent of speech \cite{modelsOfNaturalLanguageUnderstanding}. However, this strategy is not always practical because it is computationally intensive. The other primary method of intent determination involves creating a series of phrases mapped to predetermined intents. This second method (phrase mapping) is less complex and requires fewer computational resources to execute than true NLU, making it suitable for real-time intent determination. Once an NLU system has determined the user's intent, it becomes possible to translate high-level commands into a sequence of actions that can be performed by a platform.

With high-level speech commands as an input, control systems should translate wearer intents into low-level controllers, such as position, force, or impedance controllers, to synchronize gait planning in a smooth and time-continuous manner. Furthermore, high-level commands should be subject to safety constraints to avoid sudden movements, which may lead to injuries. Finite state machines (FSMs) address this motion planning problem by acting as a central planner for transitions between exoskeleton states, such as standing and walking \cite{fsm-ieee}. Thus, FSMs may serve as an interface to translate high-level user intents into motion plans for low-level controllers.

One bioinspired strategy for exoskeleton control is the central pattern generator (CPG). The CPG consists of connected nodes that can generate rhythmic patterns without receiving rhythmic inputs, facilitating joint motion synchronization necessary to recapitulate rhythmic motor behaviours such as bipedal locomotion \cite{cpg-control, legarda-bipedal-robot}. CPGs typically include parameters that allow for modulation of locomotion, which provides additional control and flexibility for the wearer \cite{autonomous-cpg, cpg-schrade}. The ability of CPGs to generate time-continuous rhythmic motions makes them an appropriate candidate for shaping the trajectories of lower limb exoskeletons, and they have been investigated in previous studies where joint synchronization was achieved through systems of coupled differential equations \cite{adaptive-cpg, cpg-gui, cpg-zhang, cpg-sharifi, zhang-neural-oscillators}.

%, such as the adjustment of gait speed,

In this study, a speech-based locomotion planning strategy combined social HRI, FSM, and CPG for the intelligent motion planning of a lower limb exoskeleton for bipedal locomotion. As opposed to a button-based interface, voice input was used as the primary mode of determining user intent, allowing hands-free use of an exoskeleton. These high-level commands were processed by an FSM to ensure safe and natural state transitions before being executed in low-level position controllers. The major contributions of the proposed control scheme can be summarized as follows: 
\begin{itemize}
    \item Integration of SR and a lower limb exoskeleton to create a system that allows a user's hands to be free to use mobility aids while still controlling the exoskeleton.
    \item The SR system used a combination of denoising, speech perception, and NLU modules to perform SR. By using gated recurrent units for denoising alongside probabilistic phrase mapping for NLU, we achieved an accurate SR system capable of running on a low-power device in real-time without requiring a remote machine to perform computation. This allows for a self-contained SR platform that can move with a user, allowing for a broader application of speech-based controls.
    \item A novel set of CPG dynamics was proposed to synchronize time-continuous transitions between exoskeleton locomotion states (e.g., sit, stand, walk) in response to discrete user inputs. Speech inputs were processed through an FSM alongside joint angles and velocities to streamline state transitions (e.g., speed up, slow down). Although voice-activated robotic systems have previously been investigated \cite{robotic-glove, wang-vc-exo}, previous CPG dynamics have not incorporated speech-based inputs for lower limb exoskeletons \cite{cpg-control}.
\end{itemize}

The proposed three-stage motion planning scheme is outlined as follows: The speech processing system is presented in Sec. \ref{sec:speech-processing}, the CPG and motion planning strategy is presented in Sec. \ref{sec:motion-planning}, the experimental evaluations of the proposed motion planning strategy are presented and discussed in Sec. \ref{sec:results}, and the concluding remarks are provided in Sec. \ref{sec:conclusion}.

\section{Speech Processing System} \label{sec:speech-processing}
\begin{figure*}[t!]
    \centering
    \includegraphics[width=\textwidth]{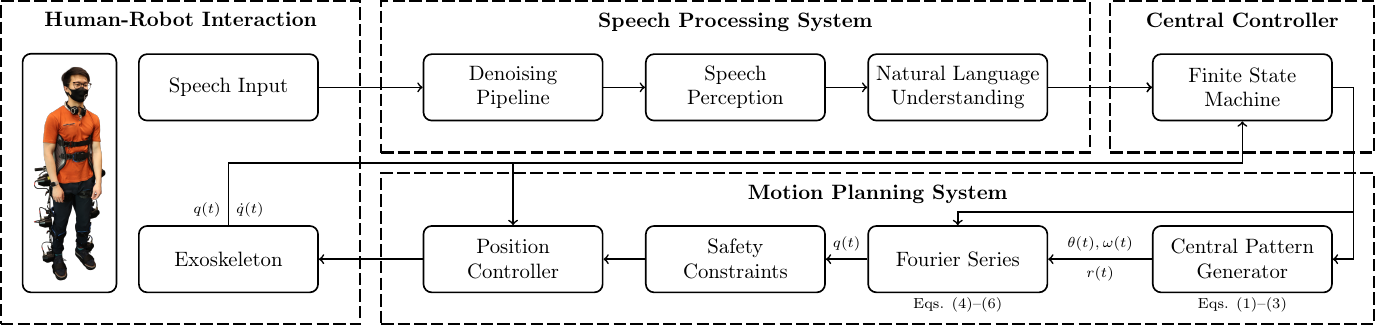}
    \caption{Proposed strategy for speech-based planning of lower limb exoskeletons.}
    \label{fig:control-strategy}
\end{figure*}

The proposed speech processing system was a customized version of the pipeline (Fig. \ref{fig:control-strategy}) developed by Reverb Robotics \cite{grasse_tata_2021_platform}, consisting of a denoising function to increase the signal-to-noise ratio (SNR), a speech perception module to process raw speech data, and a natural language understanding model to determine speech intent. Importantly, the system was able to run on a Raspberry Pi, which allowed for increased portability and for it to be applied to mobile platforms. Once an intent was determined, the Pi sent the command over a network to the central exoskeleton motion planner as shown in Fig. \ref{fig:control-strategy}. The overall system takes approximately 0.5 seconds process and send the command to the exoskeleton control system. When factoring in network latency this results in a 500-1000 ms delay between a command being spoken and actuation of the exoskeleton.

% due to its low power requirement and portability allowing it to move with the exoskeleton without impeding the user.

% \begin{figure}[t!h]
%     \centering
%     \begin{tikzpicture}[node distance=1.5cm]
%         \node (speaker) [block1] {User Speech};
%         \node (denoiser) [block1, right of=speaker, xshift=3cm] {Denoising Function};
%         \node (sp) [block1, below of=denoiser] {Speech Perception};
%         \node (nlu) [block1, below of=speaker] {Natural Language Understanding};
%         \node (output) [block1, below of=nlu] {State Change \& CPG};
%         % arrow connections
%         \draw [arrow] (speaker) -- (denoiser);
%         \draw [arrow] (denoiser) -- (sp);
%         \draw [arrow] (sp) -- (nlu);
%         \draw [arrow] (nlu) -- (output);
%     \end{tikzpicture}
%     \caption{The basic speech processing pipeline.}
%     \label{fig:speech-processing}
% \end{figure}

\subsection{Denoising Pipeline}
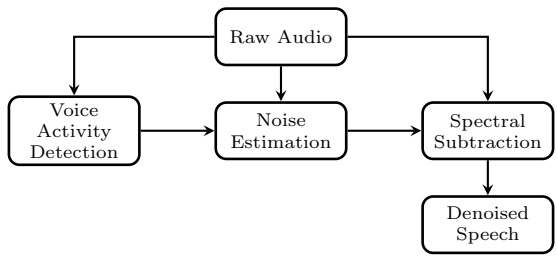
\begin{figure}[t!]
    \centering
    \begin{tikzpicture}[node distance=1.25cm]
    \tikzstyle{every node}=[font=\scriptsize]
        \node (origin) [block2] {Raw Audio};
        \node (ne) [block2, below of=origin] {Noise Estimation};
        \node (va) [block2, left of=ne, xshift=-1.5cm] {Voice Activity Detection};
        \node (ss) [block2, right of=ne, xshift=1.5cm] {Spectral Subtraction};
        \node (output) [block2, below of=ss] {Denoised Speech};
        % arrow connections
        \draw [arrow] (origin) -| (ss);
        \draw [arrow] (origin) -- (ne);
        \draw [arrow] (origin) -| (va);
        \draw [arrow] (va) -- (ne);
        \draw [arrow] (ne) -- (ss);
        \draw [arrow] (ss) -- (output);
    \end{tikzpicture}
    \caption{Major functions for denoising voice activity.}
    %\caption{The denoising pipeline. Raw Audio is the input which then filters through fucntions that detect voice activity, estimate the noise present, and then noise is subtracted from the audio and cleaned speech is output}
    \label{fig:denoising}
\end{figure}

The denoising function processed raw user inputs before passing the denoised signal to the speech perception function, increasing the SNR with the intention of improving the accuracy of the speech perception module in noisy environments. Denoising was achieved by detecting voice activity, analyzing its spectral signature to estimate noise, and subtracting its spectral signature of the noise from the input audio to produce the denoised audio (Fig. \ref{fig:denoising}). We used RNNoise, a deep recurrent neural network, to denoise the input audio for the SR system along with embeddings provided by the developers trained on 140 hours of data. RNNoise used gated recurrent units to track long-term patterns that were difficult to process with conventional recurrent units \cite{valin2018hybrid}. For example, the denoising function detected fan noise and subtracted that sound from speech. Moreover, when a user moved to a different area without that noise, the function stopped removing that spectral signature and adapted to remove any new noise.

Gated recurrent units function similarly to normal long-short term memory recurrent units with the addition of two sigmoid-activated gates. These activation gates control when the recurrent unit learns and forgets information, addressing the vanishing gradient problem with classical recurrent neural networks \cite{vanishing-gradient}. This feature of gated recurrent units allowed the denoising function to remove persistent background noise from an input signal while preserving speech information.

\subsection{Speech Perception Network}
The speech processing system uses an implementation of the Vosk speech perception system \cite{voskArchitecture} to perceive speech and form words. The Vosk implementation combines the time delay neural network factorization model and a multi-stream convolutional neural network. When the time delay neural network factorization architecture is combined with a multi-stream convolutional neural network architecture that allowed for simultaneous processing of several temporal windows, the result was a robust speech perception model. We used a pre-trained set of open-source weights called `vosk-model-small-en' provided by Alpha Cephei \cite{voskofflinespeechrecognitionapi}. Additionally, we used a constrained vocabulary to reduce the search space for SR and improve the recognition of words in the vocabulary. This is implemented by simply creating a list of words the network can perceive.

\subsection{Natural Language Understanding}
%Whereas speech perception concerns the process of parsing sound and creating words, NLU is process of understanding the meaning of speech. 
To address the problem of intent determination, the SR system used Snips NLU \cite{SnipsNLUPaper} to create a list of intents along with phrases that trigger those intents. Although a variety of phrases were associated with each intent, rather than using exact phrase checking, the intent parser used a probabilistic engine that focuses on keyword analysis, enabling the NLU engine to associate perceived speech with the proper intent without that speech being in the phrase list. Once an intent was extracted, the program determined whether the speech was directed at the exoskeleton. We checked if the command contained the word `robot' to prevent the exoskeleton from performing an unintended action. Thus, the specific key words or phrases to trigger a state transition was `robot \{keep moving, don't change, maintain speed, stand up, stand, sit down, sit, stop moving, stop, walk forward, walk, move forward, move, forward, slow down, slow, speed up, go faster, faster\},' where one of the key words or phrases in the set are used (Fig. \ref{fig:fsm}).

\section{Motion Planning and Control} \label{sec:motion-planning}

\subsection{Overview of the Exoskeleton Motion Planning System}

The proposed exoskeleton locomotion motion planning strategy combined a speech processing system, FSM, and CPG to ensure safe time-continuous transitions between exoskeleton states (Fig. \ref{fig:control-strategy}). Firstly, high-level voice commands were processed through the Reverb Robotics speech processing system \cite{grasse_tata_2021_platform}. Then, user intents were sent to the FSM of the exoskeleton. When a new intent was received, the FSM sent an input to the CPG, which fed into a Fourier series to generate joint trajectories subject to safety constraints. The period and amplitude of the trajectory could be modulated via inputs to the FSM. Then, these signals fed into position controllers, which controlled the exoskeleton. Finally, exoskeleton joint angles and velocities fed back into the FSM in a closed loop to ensure safe transitions between exoskeleton states.

\begin{table}[t!]
    \centering
    \caption{CPG parameters, initial conditions, and constants for the proposed motion planning scheme. For parameters and initial conditions, $l$ and $r$ refer to the left and right side of the exoskeleton, respectively.}
    \begin{tabular}{lll}
        \toprule
        Parameters & Initial conditions & Constants \\
        \midrule
        $v_{ij} = 0.1$ & $A_0(k) = 1$ & $c_r = 2.5$ \\
        $\phi_{ll} = \SI[parse-numbers=false]{0}{rad}$ & $\Omega_0(k) = \SI[parse-numbers=false]{\pi/2}{rad}$ & $c_\theta = \SI[parse-numbers=false]{2}{rad}$ \\
        $\phi_{rr} = \SI[parse-numbers=false]{0}{rad}$ & $\theta_l(0) = \SI[parse-numbers=false]{(2+\pi)}{rad}$ & $\beta_\omega = 10\pi$ \\
        $\phi_{lr} = \SI[parse-numbers=false]{\pi}{rad}$ & $\theta_r(0) = \SI{2}{rad}$ & $\beta_r = 10 \pi$ \\
        & & $T = \SI{2}{s}$ \\
        \bottomrule
    \end{tabular}
    \label{tab:cpg-ic-const}
\end{table}

\subsection{Synchronization of Joint Trajectories}
The proposed CPG dynamics for the phase $\theta_i(t)$ and amplitude $r(t)$ of the $i$th exoskeleton joint are governed by the following equations based on the Kuramoto model for the synchronization of coupled oscillators \cite{kuramoto}.
\begin{equation}
    \begin{split}
        \dot{\theta}_i(t) &= \omega(t) + \sum_{j=1}^{N} v_{ij} \sin(\theta_i(t) - \theta_j(t) - \phi_{ij}) \\
        \ddot\omega(t) &= \lambda(t) \beta_\omega \left( \frac{\beta_\omega}{4} ( \Omega_n(k) - \omega(t) ) - \dot\omega(t) \right) \\
        \ddot{r} (t) &= \lambda(t) \beta_r \left( \frac{\beta_r}{4} ( A_n(k) - r(t) ) - \dot{r}(t) \right)
    \end{split}
    \label{eq:cpg}
\end{equation}
where $N$ is the number of joints, $v_{ij}$ is the coupling strength, $\phi_{ij}$ is the phase offset, $\beta_\omega$ and $\beta_r$ are fixed constants, $\Omega_n(k)$ and $A_n(k)$ are user-adjustable constants which modulate the frequency and amplitude of the system, respectively, and $\lambda(t)$ is a user-triggered ramping system which multiplies the CPG signal by a linear time-dependent gain. In particular, $\Omega_n(k)$ and $A_n(k)$ update in response to user inputs, $k$ (Table \ref{tab:cpg-ic-const}).
\begin{equation}
    \begin{split}
        \Omega_n(k) &=
        \begin{cases}
            \Omega_{n-1} + c_\theta, & k=\text{speed up} \\
            \Omega_{n-1} - c_\theta, & k=\text{slow down} \\
            \Omega_{n-1}, & k=\text{otherwise}
        \end{cases} \\
        A_n(k) &=
        \begin{cases}
            A_{n-1} + c_r, & k=\text{speed up} \\
            A_{n-1} - c_r, & k=\text{slow down} \\
            A_{n-1}, & k=\text{otherwise}
        \end{cases} \\
    \end{split}
    \label{eq:speed}
\end{equation}
Here, $\Omega_n$ and $A_n$ are the updated speed constants, $\Omega_{n-1}$ and $A_{n-1}$ are the current speed constants, and $c_\theta$ and $c_r$ are constants that adjust the frequency and amplitude of the dynamics in (\ref{eq:cpg}).

The ramping system is defined as follows:
\begin{equation}
    \lambda(t) =
    \begin{cases}
        t/T, & \text{stand-to-walk} \\
        1-t/T, & \text{walk-to-stop} \\
        1, & \text{walking} \\
        0, & \text{otherwise}
    \end{cases}
    \label{eq:ramp}
\end{equation}
where the stand-to-walk and walk-to-stop conditions are triggered by user commands and held for a period $T$ (Table \ref{tab:cpg-ic-const}).

Coupling between all joints is maintained by the same principal frequency of $\omega(t)$ and amplitude $r(t)$ to synchronize locomotion trajectories. Furthermore, the coupling expression in (\ref{eq:cpg}), $v_{ij} \sin(\theta_i(t) - \theta_j(t) - \phi_{ij})$, returns the exoskeleton joints to their original trajectories if the joint phases are perturbed.

The desired walking trajectory $q_{w_i}(t)$ for the joint $i$ of the exoskeleton is defined as
\begin{equation}
    q_{w_i}(t) = r(t) \left( a_{0_i}+ \sum_{k=1}^{N_i}  (a_{k_i} \cos k\theta_i(t) + b_{k_i} \sin k\theta_i(t) ) \right)
    \label{eq:walk-traj}
\end{equation}
where $a_{k_i}$ and $b_{k_i}$ are the coefficients of Fourier series with $N_i$ terms (Table \ref{tab:fs-coefs}). The amplitude and phase of these oscillatory motions are updated in real-time by $\theta_i(t)$ and $r(t)$. Trajectories for the ankle, knee, and hip during walking were obtained from Subject 6 in the experiments of Lencioni \textit{et al.} \cite{lencioni}.

\begin{table*}[t!]
    \centering \setlength\tabcolsep{3.5pt}
    \caption{Coefficients of the Fourier series for the hip and knee from sitting, standing, and walking.$^*$}
    \scriptsize
    \begin{tabular}{p{1cm} lSSSSSSSSSSSSSSS}
        \toprule
        State & \multicolumn{1}{c}{Joint} & \multicolumn{1}{c}{$a_0$} & \multicolumn{1}{c}{$a_1$} & \multicolumn{1}{c}{$a_2$} & \multicolumn{1}{c}{$a_3$} & \multicolumn{1}{c}{$a_4$} & \multicolumn{1}{c}{$a_5$} & \multicolumn{1}{c}{$a_6$} & \multicolumn{1}{c}{$b_1$} & \multicolumn{1}{c}{$b_2$} & \multicolumn{1}{c}{$b_3$} & \multicolumn{1}{c}{$b_4$} & \multicolumn{1}{c}{$b_5$} & \multicolumn{1}{c}{$b_6$} \\
        \midrule
        \multirow{3}{1cm}{Sitting/ Standing} & Hip   & 105.40 & -1.52 & 1.29  & 0.42  & 0.36   & -0.04 & -0.12   & -3.86  & 2.92   & 0.24  & -0.49 & -0.28 & 0.01   \\
        & Knee  & 140.20 & -38.30 & -1.75 & 0.55  & 1.32   & 0.14  & n/a     & -29.22 & 8.39   & 4.14  & 0.15  & -0.39 & n/a    \\
        & Ankle & 49.59 & 22.01 & -1.31 & 0.28  & -0.07 & 0.15  & n/a     & 35.11  & -11.48 & -4.05 & -0.16 & 0.40   & n/a    \\
        \midrule
        \multirow{3}{*}{Walk} & Hip   & 40.69 & 23.22 & -4.49 & 0.40  & 0.70   & 1.08  & -0.27   & -8.65  & 3.34   & 1.39  & 0.80   & 0.34  & 0.07   \\
        & Knee  & 25.70  & -3.83 & -8.54 & 1.91  & 1.09   & 2.05  & -0.31   & -19.28 & 17.93  & 3.77  & 1.50   & 0.58  & -0.90   \\
        & Ankle & -0.99 & 5.29  & -8.58 & -0.48 & 1.69   & -0.04 & 1.30    & 3.61   & -5.95  & 5.92  & -2.12 & 1.02  & -0.72  \\
        \bottomrule \\[-6pt]
        \multicolumn{10}{l}{\scriptsize $^{*}R^2_{\text{adj}} \geq 0.99$ for each trajectory.}
    \end{tabular}
    \label{tab:fs-coefs}
\end{table*}

% \subsection{Sit-to-Stand and Stand-to-Sit Trajectories}
% \begin{figure}[t!]
%     \centering
%     \begin{tikzpicture}
%         \begin{axis}[
%             xlabel=Movement Percent (\%),
%             ylabel=Angle ($^\circ$),
%             no markers,
%             every axis plot/.append style={thick},
%             legend cell align={left},
%         ]
%             \addplot [blue] table [x=x1, y=ankle_traj, col sep=comma] {\fourierdata};
%             \addplot [black] table [x=x1, y=knee_traj, col sep=comma] {\fourierdata};
%             \addplot [red] table [x=x1, y=hip_traj, col sep=comma] {\fourierdata};
%             \legend{Ankle, Knee, Hip};
%         \end{axis}
%     \end{tikzpicture}
%     \caption{Fourier series for the sit-to-stand trajectory of the ankle, knee, and hip (equation \ref{eq:sit_stand_exo_coords}). Data obtained from the experiments of Nuzik et al. \cite{nuzik}.}
%     \label{fig:fourier_trajs}
% \end{figure}

The sit-to-stand and stand-to-sit trajectories for the ankle, knee, and hip are obtained from the experiments of Nuzik \textit{et al.} \cite{nuzik}, which included motion data for 55 healthy adults (38 women, 17 men). The mean trajectory values of all subjects for each joint were used to compute the desired sit-to-stand and stand-to-sit trajectory $q_{s_i}(t)$ for the joint $i$ of the exoskeleton.
\begin{equation}
    q_{s_i}(t) = a_{0_i} + \sum_{k=1}^{N_i} (a_{k_i} \cos k \omega_{s_i} t + b_{k_i} \sin k \omega_{s_i} t)
    \label{eq:fourier_sit_stand}
\end{equation}
where $\omega_{s_i}$ is the angular frequency of the trajectory, and $a_{k_i}$ and $b_{k_i}$ are the coefficients of the Fourier series with $N_i$ terms (Table \ref{tab:fs-coefs}). Equation (\ref{eq:fourier_sit_stand}) was set to satisfy the time-dependent boundary conditions $\dot{q}_{s_i}(t_0) = 0$ and $\dot{q}_{s_i}(T) = 0$, where $t_0$ is the initial time and $T$ is the time after one sit-to-stand or stand-to-sit period (Table \ref{tab:cpg-ic-const}). The output of (\ref{eq:fourier_sit_stand}) was transformed into the coordinate system of the exoskeleton via the linear function
\begin{equation}
    q_{t,s_i}(t) = -q_{s_i}(t) + \max (q_{s_i}(t))
    \label{eq:sit_stand_exo_coords}
\end{equation}
where $q_{t, s_i}(t)$ is the transformed angle. This function ensures that the endpoint for the sit-to-stand trajectory for the ankle, knee, and hip terminates at $0^\circ$. The stand-to-sit trajectory is implemented as the reverse of the sit-to-stand trajectory (i.e., $q_{t,s_i}(t) \mapsto q_{t,s_i}(-t)$ in equation \ref{eq:sit_stand_exo_coords}).

\subsection{Safety Considerations}

\begin{figure}[t!]
    \centering
    \begin{tikzpicture}[node distance=2.25cm, on grid, auto,
    state/.style={circle, draw, minimum size=0.7cm, text width=0.7cm, align=center, initial text=},
    every initial by arrow/.style={*->, >=stealth, thick}]
    \tikzstyle{every node}=[font=\scriptsize]
        % states
        \node[state] (sit) {Sit};
        \node[state, right of=sit] (stand) {Stand};
        \node[state, above right of=stand, minimum size=1.2cm, text width=1.2cm] (rup) {Locomotion Initiation};
        \node[state, below right of=stand, minimum size=1.2cm, text width=1.2cm] (rdown) {Locomotion Completion};
        \node[state, below right of=rup] (walk) {Walk};
        % state transitions
        \draw[arrow] (sit) edge [bend left, above] node {stand} (stand)
                     (stand) edge [bend left, below] node {sit} (sit)
                     % ramp up
                     (rup) edge (walk)
                     (stand) edge [below right] node [xshift=-.5em] {walk} (rup)
                     % ramp down
                     (walk) edge [above left] node [xshift=.5em] {stop} (rdown)
                     (rdown) edge (stand)
                     % walk states
                     (walk) edge [loop above, above] node {speed up} (walk)
                     (walk) edge [loop below, below] node {slow down} (walk);
    \end{tikzpicture}
    \caption{Proposed FSM to plan the transitions between exoskeleton states.}
    \label{fig:fsm}
\end{figure}
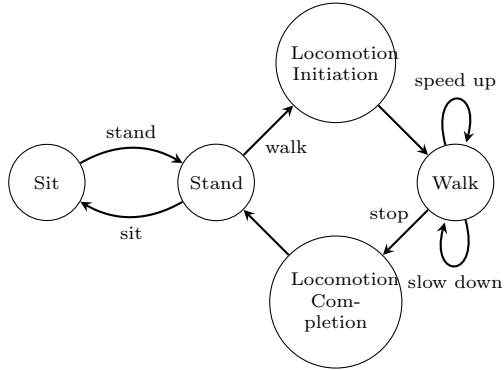

The finite state machine was designed to ensure safe transitions between exoskeleton states. The states and intents for triggering transition can be seen in figure \ref{fig:fsm}, and the initial state of the exoskeleton is either sitting or standing, depending on the initial position of the user. Position controller constraints include limitations on the maximum torque and velocity and maximum and minimum joint angles.

% The five exoskeleton states are sit, stand, locomotion initiation, locomotion completion, and walk. Firstly, the walk command transitions the exoskeleton from stand to locomotion initiation to walk in one continuous motion. Secondly, the stop command transitions the exoskeleton from walk to locomotion completion to stand in one continuous motion. Thirdly, the ramp ensures a smooth transition between standing and walking and vice versa. Finally, the commands speed up and slow down modulate the walking speed of the exoskeleton only during the walk state. While walking, the exoskeleton does not sit, stand, or restart the walking trajectory. 

\section{Experimental Evaluations} \label{sec:results}

\subsection{Speech Processing Experiments}
%experiment design flaws: didn't use a standard mask across participants.
Experimental assessments were performed to evaluate the efficacy of the SR system with ethics permission from the University of Lethbridge Human Subjects Review Board, number 2013-037. In the experiment, participants removed their protective mask being worn due to public health restrictions and spoke a series of pre-defined commands consistent with what would be used in normal use of the exoskeleton. After each command, the output of the SR system and NLU engine was recorded. Then, the participant put on their mask and repeated the experiment. SR performance was measured with two metrics, word error rate (WER) and intent error rate (IER). WER is a standard metric for speech perception performance that is determined by the ratio of insertions ($I$), substitutions ($S$), and deletions ($D$) required to transform a response into the target to the number of words in the target $N$:
\begin{equation}
    \text{WER} = \frac{I + S + D}{N} \times 100\%
    \label{eq:WER}
\end{equation}
IER is a measure of whether the correct intent was derived from the user's speech:
\begin{equation}
    \text{IER} = \frac{1}{N} \left( \sum_{\text{all trials}} \begin{cases}
        0, & \text{correct output} \\
        1, & \text{incorrect output}
    \end{cases} \right) \times 100\%
    \label{eq:IER}
\end{equation}

This metric has a similar function to the Stanford Sentiment Treebank (SST) metric\cite{SST} that is a part of a standard NLU testing corpus General Language Understanding Evaluation\cite{GLUE}.  SST describes the accuracy of sentiment analysis in a single sentence, though given the highly constrained nature of the model used in our project, directly using SST would not provide an accurate assessment of this model. Therefore IER was used for intent accuracy analysis. After WER and IER were calculated for individual trials, the results were aggregated to get a cumulative average for both metrics.

% \begin{table}[t!]
%     \centering \setlength\tabcolsep{4pt}
%     \caption{Results for human SR trials.}
%     \begin{tabular}{p{1.75cm} lSS}
%         \toprule
%         Mask State & \multicolumn{1}{c}{Demographic} & \multicolumn{1}{c}{WER} & \multicolumn{1}{c}{IER} \\
%         \midrule
%         \multirow{3}{*}{Unmasked} & Female & 12.11\% & 14.29\% \\
%         & Male   & 9.30\% & 10.71\% \\
%         & Combined & 10.50\% & 12.24\% \\
%         \midrule
%         \multirow{3}{*}{Masked} & Female  & 13.46\%  & 11.11\% \\
%         & Male   & 7.52\% & 13.10\% \\
%         & Combined & 10.07\% & 12.24\%  \\
%         \midrule
%         \multirow{3}{*}{Combined} & Female  & 12.78\%  & 12.70\% \\
%         & Male   & 8.41\% & 11.52\% \\
%         & Combined & 10.29\% & 12.24\%  \\
%         \bottomrule
%     \end{tabular}
%     \label{tab:SR-analysis}
% \end{table}

\begin{figure}[t!]
    \centering
    \begin{subfigure}{0.475\textwidth}
        \centering
        \includegraphics[width=\columnwidth]{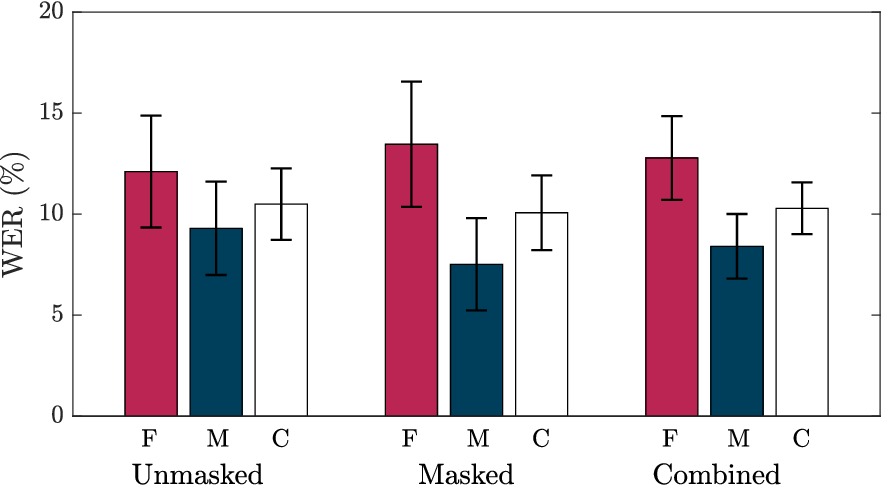}
        \vspace*{.5mm} \hspace*{2em}
        \caption{}
    \end{subfigure}%
    \hfill
    \begin{subfigure}{0.475\textwidth}
        \centering
        \includegraphics[width=\columnwidth]{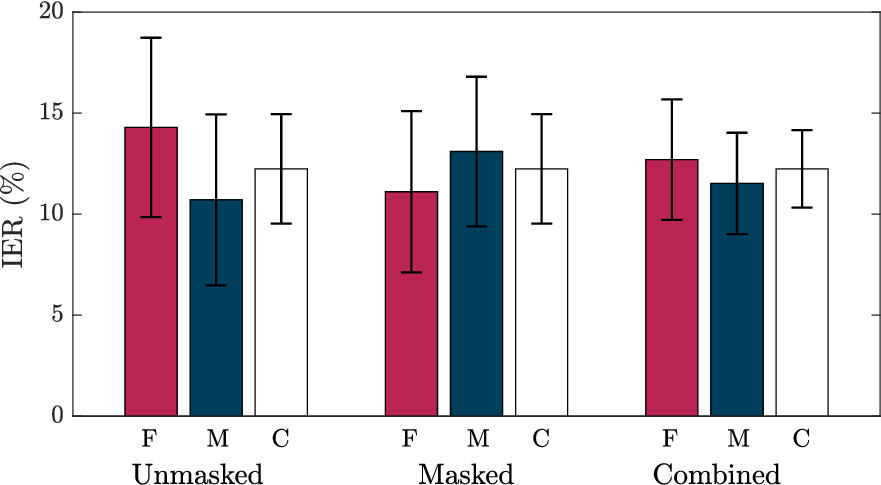}
        \vspace*{.5mm} \hspace*{2em}
        \caption{}
    \end{subfigure}
    \caption{Results for the human SR trials for (a) WER and (b) IER. F$=$Female, M$=$Male, C$=$Combined. There were no significant differences between or within groups.}
    \label{fig:SR-graphs}
\end{figure}

Experiments were performed by 4 male and 3 female participants. Participants were taken from a population of undergraduate and graduate students at the University of Lethbridge aged between 19 and 31. WER and IER scores from each participant were aggregated and computed (Fig. \ref{fig:SR-graphs}).

%  Notably, because the experiment was designed only to evaluate the SR system, we did not reject participants or exclude their results based on their first language. 

One drawback of the SR system has to do with the intent recognition system. By implementing a constrained vocabulary, words spoken by the user may be similar enough to words in the vocabulary where an intent might be triggered. For example, `hello' may register as `slow'. We reduced this problem by checking for the keyword `robot' before intent determination. In locomotion experiments, this feature resulted in the robot not executing commands when the word `robot' was missed, particularly when a user moved into a new acoustic environment (participants moved between a quiet lab area and a room with fume hood noise) where the denoising function was not updating its model fast enough.

% A constrained vocabulary, as implemented in our SR system, could decrease the WER when perceiving speech by removing words that could be misunderstood from the pool of perceivable words. Moreover, the use of probabilistic phrase mapping could result in a low IER. One limitation of a 

\subsection{Locomotion Tasks With and Without Speech Commands}

\begin{figure}[t!]
    \centering
    \begin{subfigure}{0.15\columnwidth}
        \centering
        \includegraphics[width=\textwidth]{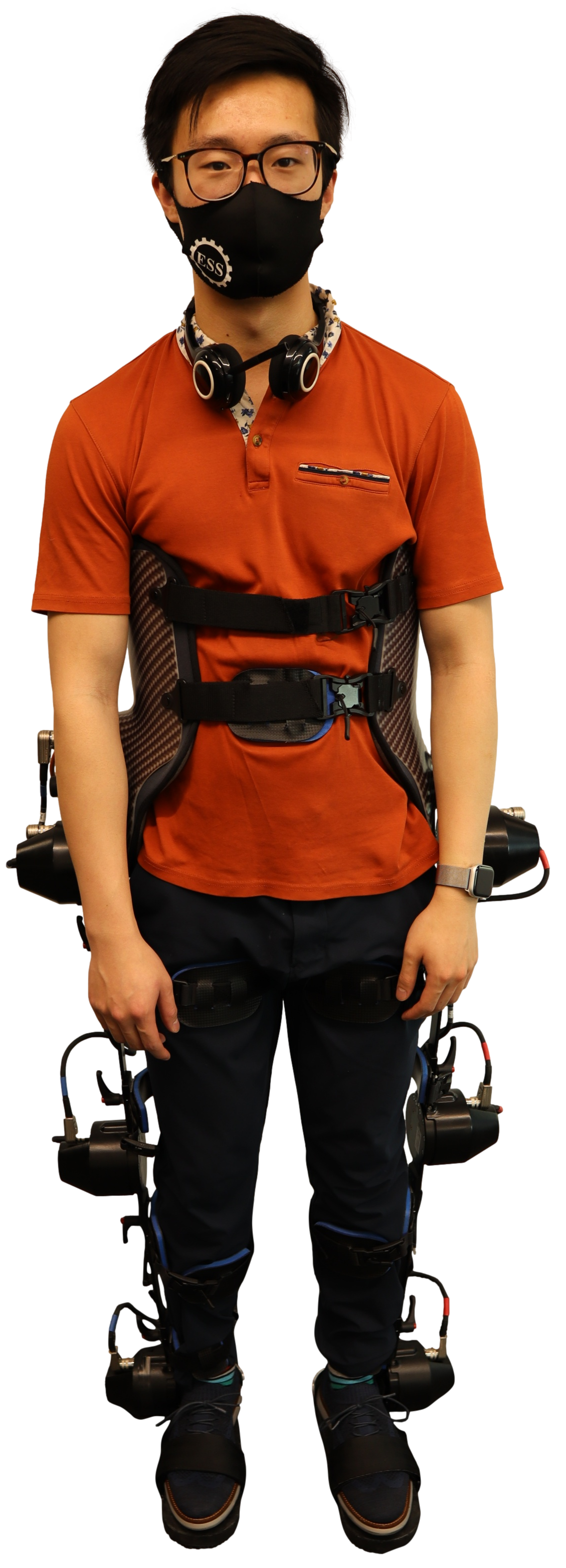}
        \caption{}
        \label{fig:eddie-exo}
    \end{subfigure}
    \hspace{5em}
    \begin{subfigure}{0.15\columnwidth}
        \centering
        \includegraphics[width=\textwidth]{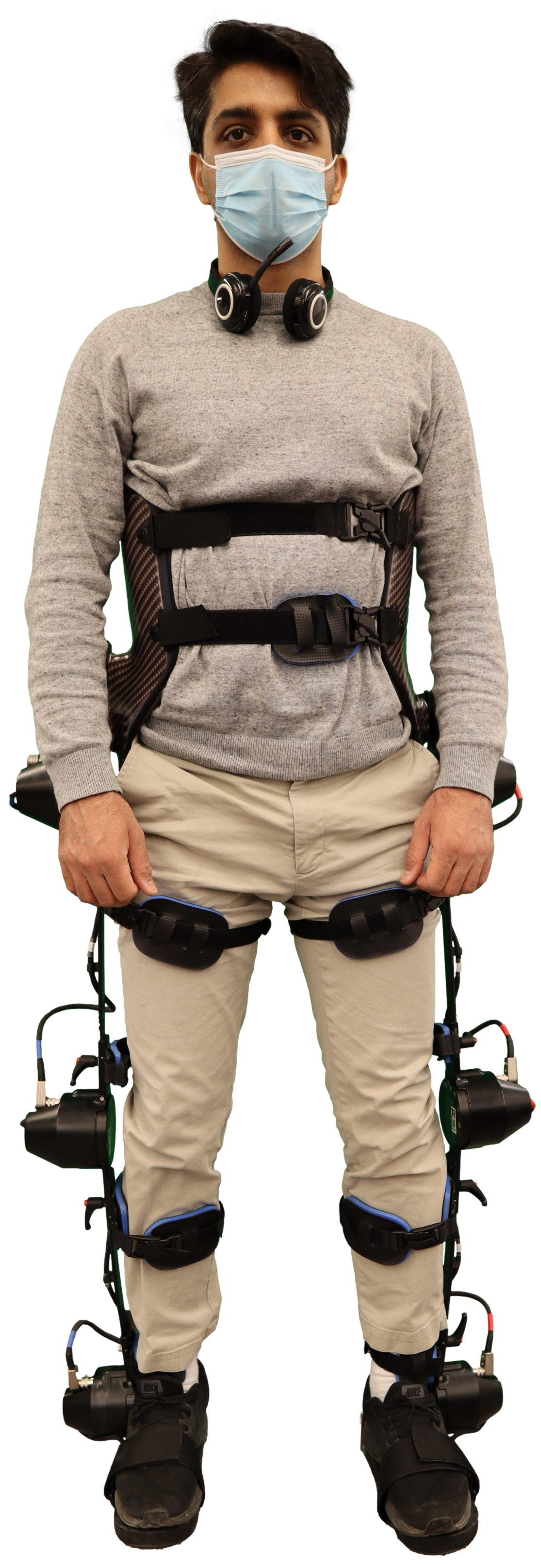}
        \caption{}
        \label{fig:mojtaba-exo}
    \end{subfigure}
    \caption{Exo-H3 lower limb exoskeleton worn by two able-bodied users for walking. The participants use a wireless headset to command the exoskeleton. (a) User 1 (21-year-old) and (b) user 2 (33-year-old).}
    \label{fig:exo-wearers}
\end{figure}

The proposed motion planning strategy was assessed experimentally to provide proof-of-concept evidence for the effectiveness of speech-based locomotion planning with the Exo-H3 lower limb exoskeleton from Technaid. The proposed motion planning scheme was implemented in real-time using MATLAB Simulink which received sensory data and controlled motors at a sampling rate of \SI{100}{\hertz} via a CAN interface (Vector VN1610) with 2 channels. The SR system was run on a Raspberry Pi 3, which received speech input from the microphone on a Logitech Wireless Headset H600 at a sampling rate of \SI{16000}{\hertz}, and the processed signals were transmitted to the laptop via the user datagram protocol. The major computations in this strategy involve time integration of the CPG dynamics in equation (\ref{eq:cpg}) and calculating each joint angle for the exoskeleton with Fourier series in equations (\ref{eq:walk-traj}), (\ref{eq:fourier_sit_stand}), and (\ref{eq:sit_stand_exo_coords}). Preliminary tests were performed in a trial-and-error manner to obtain reasonable parameters for speed modulation ($c_\theta$ and $c_r$ in equation \ref{eq:speed}) and initial values and parameters of the proposed CPG system (Table \ref{tab:cpg-ic-const}). In the experiments, two able-bodied users wore the exoskeleton using a walker for postural stability (Fig. \ref{fig:exo-wearers}).

% on a laptop with an Intel Core i7-8650U CPU at 1.90 GHz and 8 GB of RAM,

\begin{figure}[t!]
    \centering
    \begin{subfigure}{\columnwidth}
        \centering
        \includegraphics[width=0.65\columnwidth]{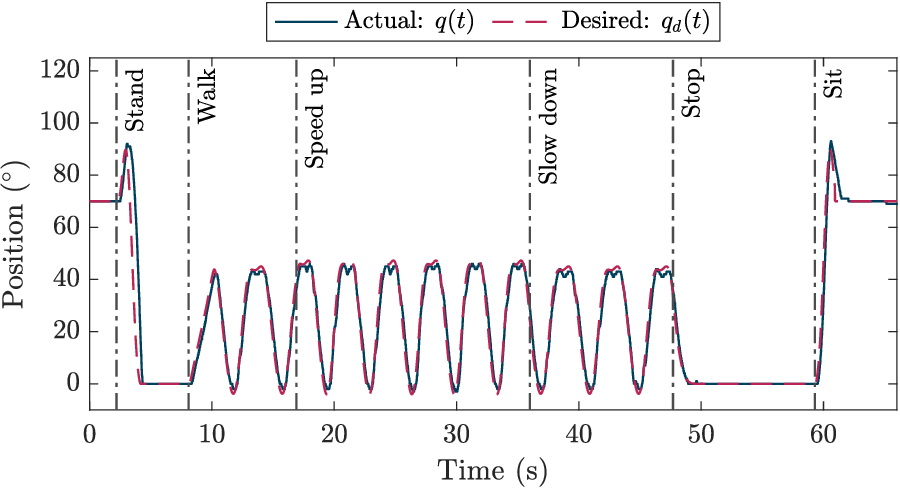}
        \hspace*{1em}
        % \caption{Left Hip}
        \caption{}
    \end{subfigure}
    \par\smallskip
    \begin{subfigure}{\columnwidth}
        \centering
        \includegraphics[width=0.65\columnwidth]{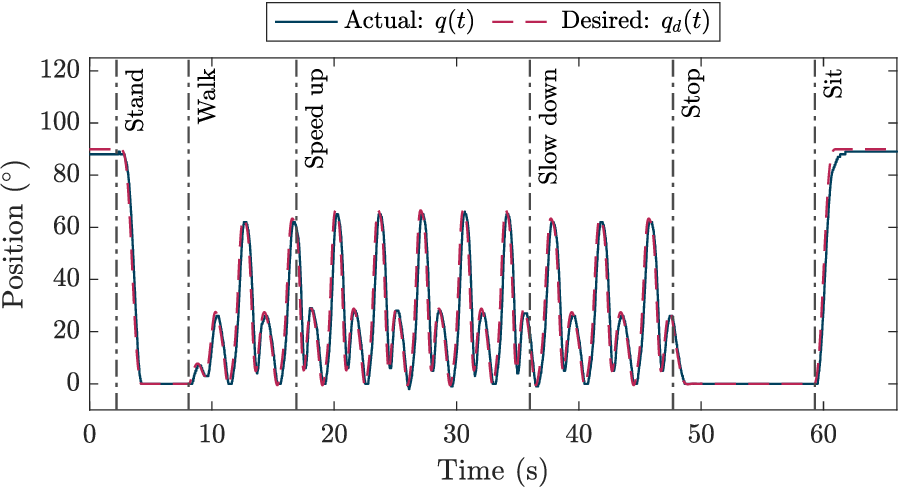}
        \hspace*{1em}
        % \caption{Left Knee}
        \caption{}
    \end{subfigure}
    \par\smallskip
    \begin{subfigure}{\columnwidth}
        \centering
        \includegraphics[width=0.65\columnwidth]{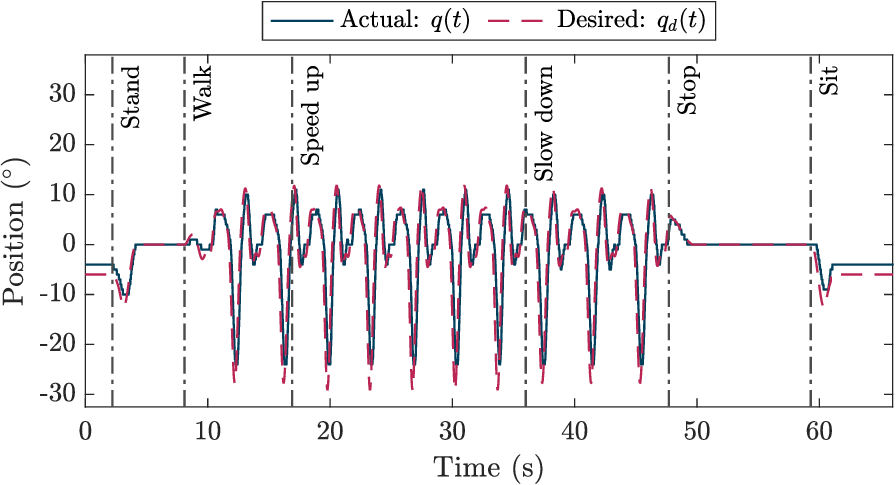}
        \hspace*{1em}
        % \caption{Left Ankle}
        \caption{}
    \end{subfigure}
    \caption{Actual and desired trajectories for the left (a) hip, (b) knee, and (c) ankle for the A to B locomotion task for user 1. Vertical lines correspond to the labelled speech inputs.}
    \label{fig:angles}
\end{figure}

\begin{figure}[t!]
    \centering
    \includegraphics[width=0.65\columnwidth]{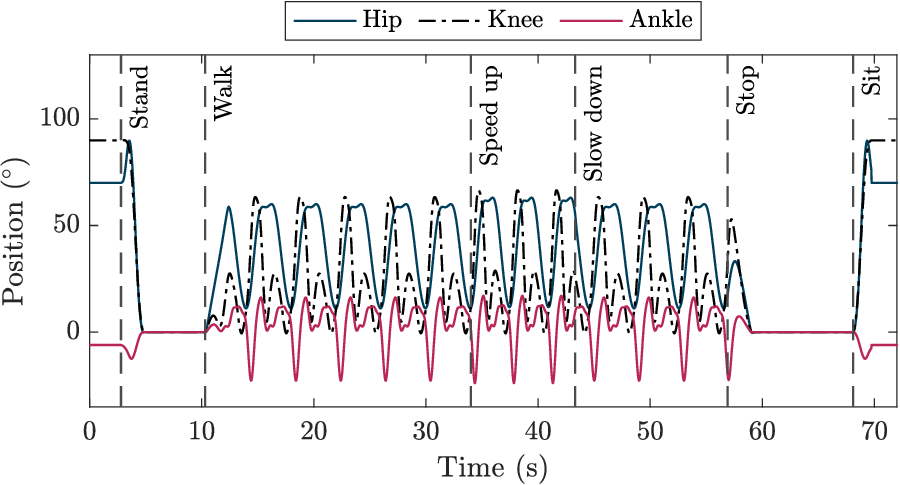}
    \caption{Desired trajectories for the left hip, knee, and ankle for the A to B locomotion task for user 2. Vertical lines correspond to the labeled speech inputs.}
    \label{fig:angle-moj}
\end{figure}

\begin{figure}[t!]
    \centering
    \includegraphics[width=0.6\columnwidth]{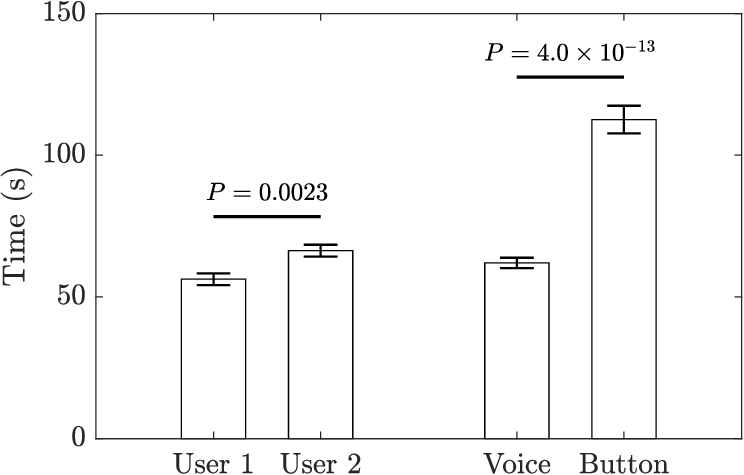}
    \caption{Time taken by users 1 and 2 to complete a locomotion task using voice control ($n_1 = 9,\ n_2 = 12$). Voice is the weighted mean time of users 1 and 2 to complete the task with voice control. Button is the mean time of user 1 to complete the task with a button-based interface ($n=10$).}
    \label{fig:stats}
\end{figure}

The first experiment involved walking from one location, A, to another location, B, in a straight line for approximately \SI{12}{\metre}. The participants began sitting at point A, then stood and walked to point B, where they sat. Between points A and B, the user could choose to speed up or slow down ad-lib. A 21-year-old participant, user 1, and a 33-year-old participant, user 2, completed 9 and 12 trials, respectively. The time to walk in these trials for user 1 was \SI{56(2)}{\second} and that for user 2 was \SI{66(2)}{\second} ($P = 0.0023$ for a two-tailed t-test assuming equal variances, Fig. \ref{fig:stats}). The significant difference in time between users 1 and 2 to complete this experiment can be attributed to user preferences in walking speeds. The trajectory for walking from point A to point B for one representative trial of user 1 is plotted in Fig. \ref{fig:angles} and that for user 2 is plotted in Fig. \ref{fig:angle-moj}. These figures show the response of the exoskeleton to voice commands.

The same experiment from A to B was performed using a button-based smartphone app from Technaid with user 1 as the participant for 10 trials, and the time taken to walk from A to B was \SI{113(5)}{\second}. In comparison with the voice command time of \SI{62(2)}{\second}, a two-tailed t-test assuming equal variances yielded $P = 4.0 \times 10^{-13}$ (Fig. \ref{fig:stats}). This significant difference can be attributed to the additional time the user needed to stop the exoskeleton, remove their hands from their walker, search for the appropriate command on the exoskeleton remote control, and press that button.

% (instead of voice commands)

\section{Conclusions} \label{sec:discussion}

In this study, a speech-based locomotion planning strategy was developed and tested to provide safe and intuitive motion planning of a lower limb exoskeleton. The proposed strategy combined a speech processing system, FSM, and CPG to plan the exoskeleton movements based on user intents.  Furthermore, a novel set of speech-based CPG dynamics was developed to synchronize exoskeleton joints in a time-continuous manner in response to discrete user inputs.

% Within the speech processing system, the employed NLU uses probabilistic phrase mapping to determine user intents, as it is less computationally intense than true NLU systems, making it suitable for real-time SR on the limited power hardware used.

To test the performance of the proposed speech processing system, WER and IER were computed as 10.29\% and 12.24\%, respectively, using speech data from 4 male and 3 female participants aged between 19 and 31 under both masked and unmasked conditions. The IER demonstrated is relatively low when considering limitations imposed by the hardware, as desktop computer-executed models have shown 9.6-4.4\% error rates in similar tests \cite{nlu-compare}, though there is certainly room for improvement. Overall the proposed SR system could be a practical option for speech-based HRI applications when considering its performance in constrained tasks and hardware it is capable of running on.

In the user studies with the exoskeleton, voice control was compared with button control in a point-to-point locomotion task. The completion time using voice commands was 54\% faster than employing a button-based mobile app interface, suggesting that voice commands may be more efficient than button-based interfaces. In addition to being efficient, speech-based planning allowed users to have both hands free to enhance postural stability and support. Thus, the combination of safety and efficiency of speech-based planning suggests it may be a promising candidate to plan the motion of lower limb exoskeletons.

% Future studies will expand this proof-of-concept speech-based motion planning strategy through larger randomized control trials of button-based vs speech-based control and by integrating personalized locomotion trajectories for user comfort.

\subsection{Future Directions}
The results of this research provide direction for future studies. First, we intend to test this system with individuals with disabilities as all participants were able-bodied. Second, we would like to improve the denoising function so it can better compensate for dynamic background noise. Finally, we want to expand the FSM and implement personalized locomotion trajectories to improve the real-world viability of the system.

\section*{CRediT Authorship Contribution Statement}

\textbf{Eddie Guo:} Conceptualization, methodology, software, experiments, and writing. \textbf{Chris Perlette:} Methodology, software, experiments, and writing. \textbf{Mojtaba Sharifi:} Conceptualization, methodology, experiments, writing, supervision, review, and editing. \textbf{Lukas Grasse:} Software. \textbf{Matthew Tata:} Supervision, review, and editing. \textbf{Vivan K. Mushahwar:} Supervision, review, and editing. \textbf{Mahdi Tavakoli:} Supervision, review, and editing.

\section*{Declaration of Competing Interests}

Authors Matthew Tata and Lukas Grasse own Reverb Robotics and developed the ReRo-Core software used as the basis for the speech recognition system.

\section*{Funding}

This work was supported by the Natural Sciences and Engineering Research Council (NSERC), Canada Foundation for Innovation (CFI), Alberta Jobs, Economy and Innovation Ministry’s Major Initiatives Fund to the Center for Autonomous Systems in Strengthening Future Communities, and Autonomous Systems Initiative (ASI).

\bibliographystyle{IEEEtran}
\bibliography{refs}

\begin{thebibliography}{10}
\providecommand{\url}[1]{#1}
\csname url@samestyle\endcsname
\providecommand{\newblock}{\relax}
\providecommand{\bibinfo}[2]{#2}
\providecommand{\BIBentrySTDinterwordspacing}{\spaceskip=0pt\relax}
\providecommand{\BIBentryALTinterwordstretchfactor}{4}
\providecommand{\BIBentryALTinterwordspacing}{\spaceskip=\fontdimen2\font plus
\BIBentryALTinterwordstretchfactor\fontdimen3\font minus
  \fontdimen4\font\relax}
\providecommand{\BIBforeignlanguage}[2]{{%
\expandafter\ifx\csname l@#1\endcsname\relax
\typeout{** WARNING: IEEEtran.bst: No hyphenation pattern has been}%
\typeout{** loaded for the language `#1'. Using the pattern for}%
\typeout{** the default language instead.}%
\else
\language=\csname l@#1\endcsname
\fi
#2}}
\providecommand{\BIBdecl}{\relax}
\BIBdecl

\bibitem{neuro-impair}
A.~Rodríguez-Fernández, J.~Lobo-Prat, and J.~M. Font-Llagunes, ``Systematic
  review on wearable lower-limb exoskeletons for gait training in neuromuscular
  impairments,'' \emph{Journal of NeuroEngineering and Rehabilitation},
  vol.~18, 2021.

\bibitem{rewalk}
G.~Zeilig, H.~Weingarden, M.~Zwecker, I.~Dudkiewicz, A.~Bloch, and
  A.~Esquenazi, ``Safety and tolerance of the rewalk™ exoskeleton suit for
  ambulation by people with complete spinal cord injury: A pilot study,''
  \emph{The Journal of Spinal Cord Medicine}, vol.~35, no.~2, pp. 96--101,
  2012.

\bibitem{indego}
S.~A. Murray, R.~J. Farris, M.~Golfarb, C.~Hartigan, C.~Kandilakis, and
  D.~Truex, ``Fes coupled with a powered exoskeleton for cooperative muscle
  contribution in persons with paraplegia,'' in \emph{2018 40th Annual
  International Conference of the IEEE Engineering in Medicine and Biology
  Society (EMBC)}, 2018, pp. 2788--2792.

\bibitem{hal}
O.~Jansen, D.~Grasmuecke, R.~C. Meindl, M.~Tegenthoff, P.~Schwenkreis,
  M.~Sczesny-Kaiser, M.~Wessling, T.~A. Schildhauer, C.~Fisahn, and M.~Aach,
  ``Hybrid assistive limb exoskeleton hal in the rehabilitation of chronic
  spinal cord injury: Proof of concept; the results in 21 patients,''
  \emph{World Neurosurgery}, vol. 110, pp. e73--e78, 2018.

\bibitem{exo-h3}
K.~A. Inkol and J.~McPhee, ``Assessing control of fixed-support balance
  recovery in wearable lower-limb exoskeletons using multibody dynamic
  modelling,'' in \emph{2020 8th IEEE RAS/EMBS International Conference for
  Biomedical Robotics and Biomechatronics (BioRob)}, 2020, pp. 54--60.

\bibitem{VILLANI2018}
\BIBentryALTinterwordspacing
V.~Villani, F.~Pini, F.~Leali, and C.~Secchi, ``Survey on human–robot
  collaboration in industrial settings: Safety, intuitive interfaces and
  applications,'' \emph{Mechatronics}, vol.~55, pp. 248--266, 2018. [Online].
  Available:
  \url{https://www.sciencedirect.com/science/article/pii/S0957415818300321}
\BIBentrySTDinterwordspacing

\bibitem{human-robotInteractionStatus}
T.~B. Sheridan, ``Human–robot interaction: Status and challenges,''
  \emph{Human Factors}, vol.~58, no.~4, pp. 525--532, 2016.

\bibitem{humanHRI}
J.~Kennedy, S.~Lemaignan, C.~Montassier, P.~Lavalade, B.~Irfan,
  F.~Papadopoulos, E.~Senft, and T.~Belpaeme, ``Child speech recognition in
  human-robot interaction: Evaluations and recommendations,'' in
  \emph{Proceedings of the 2017 ACM/IEEE International Conference on
  Human-Robot Interaction}, ser. HRI '17.\hskip 1em plus 0.5em minus
  0.4em\relax New York, NY, USA: Association for Computing Machinery, 2017, p.
  82–90.

\bibitem{speech-recognition-status}
\BIBentryALTinterwordspacing
X.~Huang, J.~Baker, and R.~Reddy, ``A historical perspective of speech
  recognition,'' \emph{Commun. ACM}, vol.~57, no.~1, p. 94–103, jan 2014.
  [Online]. Available: \url{https://doi.org/10.1145/2500887}
\BIBentrySTDinterwordspacing

\bibitem{TDNNFEfficacy}
T.~F. Abidin, A.~Misbullah, R.~Ferdhiana, M.~Z. Aksana, and L.~Farsiah, ``Deep
  neural network for automatic speech recognition from indonesian audio using
  several lexicon types,'' in \emph{2020 International Conference on Electrical
  Engineering and Informatics (ICELTICs)}, 2020, pp. 1--5.

\bibitem{moritz2016acoustic}
N.~Moritz, J.~Schr{\"o}der, S.~Goetze, J.~Anem{\"u}ller, and B.~Kollmeier,
  ``Acoustic scene classification using time-delay neural networks and
  amplitude modulation filter bank features,'' \emph{complexity}, vol.~12,
  p.~13, 2016.

\bibitem{modelsOfNaturalLanguageUnderstanding}
M.~Bates, ``Models of natural language understanding,'' \emph{Proceedings of
  the National Academy of Sciences}, vol.~92, no.~22, pp. 9977--9982, 1995.

\bibitem{fsm-ieee}
K.~A. Strausser and H.~Kazerooni, ``The development and testing of a human
  machine interface for a mobile medical exoskeleton,'' in \emph{2011 IEEE/RSJ
  International Conference on Intelligent Robots and Systems}, 2011, pp.
  4911--4916.

\bibitem{cpg-control}
A.~J. Ijspeert, ``Central pattern generators for locomotion control in animals
  and robots: A review,'' \emph{Neural Networks}, vol.~21, no.~4, pp. 642--653,
  2008, robotics and Neuroscience.

\bibitem{legarda-bipedal-robot}
\BIBentryALTinterwordspacing
J.~Arcos-Legarda, J.~Cortes-Romero, and A.~Tovar, ``Robust compound control of
  dynamic bipedal robots,'' \emph{Mechatronics}, vol.~59, pp. 154--167, 2019.
  [Online]. Available:
  \url{https://www.sciencedirect.com/science/article/pii/S095741581930039X}
\BIBentrySTDinterwordspacing

\bibitem{autonomous-cpg}
M.~Sharifi, J.~K. Mehr, V.~K. Mushahwar, and M.~Tavakoli, ``Autonomous
  locomotion trajectory shaping and nonlinear control for lower-limb
  exoskeletons,'' \emph{IEEE/ASME Transactions on Mechatronics}, vol.~27,
  no.~2, pp. 645--655, 2022.

\bibitem{cpg-schrade}
S.~O. Schrade, Y.~Nager, A.~R. Wu, R.~Gassert, and A.~Ijspeert, ``Bio-inspired
  control of joint torque and knee stiffness in a robotic lower limb
  exoskeleton using a central pattern generator,'' in \emph{2017 International
  Conference on Rehabilitation Robotics (ICORR)}, 2017, pp. 1387--1394.

\bibitem{adaptive-cpg}
M.~Sharifi, J.~K. Mehr, V.~K. Mushahwar, and M.~Tavakoli, ``Adaptive cpg-based
  gait planning with learning-based torque estimation and control for
  exoskeletons,'' \emph{IEEE Robotics and Automation Letters}, vol.~6, no.~4,
  pp. 8261--8268, 2021.

\bibitem{cpg-gui}
K.~Gui, H.~Liu, and D.~Zhang, ``A generalized framework to achieve coordinated
  admittance control for multi-joint lower limb robotic exoskeleton,'' in
  \emph{2017 International Conference on Rehabilitation Robotics (ICORR)},
  2017, pp. 228--233.

\bibitem{cpg-zhang}
D.~Zhang, Y.~Ren, K.~Gui, J.~Jia, and W.~Xu, ``Cooperative control for a hybrid
  rehabilitation system combining functional electrical stimulation and robotic
  exoskeleton,'' \emph{Frontiers in Neuroscience}, vol.~11, p. 725, 2017.

\bibitem{cpg-sharifi}
J.~K. Mehr, M.~Sharifi, V.~K. Mushahwar, and M.~Tavakoli, ``Intelligent
  locomotion planning with enhanced postural stability for lower-limb
  exoskeletons,'' \emph{IEEE Robotics and Automation Letters}, vol.~6, no.~4,
  pp. 7588--7595, 2021.

\bibitem{zhang-neural-oscillators}
\BIBentryALTinterwordspacing
X.~Zhang and M.~Hashimoto, ``Synchronization-based trajectory generation method
  for a robotic suit using neural oscillators for hip joint support in
  walking,'' \emph{Mechatronics}, vol.~22, no.~1, pp. 33--44, 2012. [Online].
  Available:
  \url{https://www.sciencedirect.com/science/article/pii/S0957415811001723}
\BIBentrySTDinterwordspacing

\bibitem{robotic-glove}
\BIBentryALTinterwordspacing
Y.~Guo, W.~Xu, S.~Pradhan, C.~Bravo, and P.~Ben-Tzvi, ``Personalized voice
  activated grasping system for a robotic exoskeleton glove,''
  \emph{Mechatronics}, vol.~83, p. 102745, 2022. [Online]. Available:
  \url{https://www.sciencedirect.com/science/article/pii/S0957415822000058}
\BIBentrySTDinterwordspacing

\bibitem{wang-vc-exo}
X.~Wang, P.~Tran, S.~Callahan, S.~Wolf, and J.~Desai, ``Towards the development
  of a voice-controlled exoskeleton system for restoring hand function,'' 04
  2019, pp. 1--7.

\bibitem{grasse_tata_2021_platform}
L.~S. Grasse and M.~S. Tata, ``An end-to-end platform for human-robot speech
  interaction,'' \emph{Sound in HRI Workshop. 16th Annual Conference for Basic
  and Applied Human-Robot Interaction.}, Mar 2021.

\bibitem{valin2018hybrid}
J.-M. Valin, ``A hybrid dsp/deep learning approach to real-time full-band
  speech enhancement,'' 2018.

\bibitem{vanishing-gradient}
\BIBentryALTinterwordspacing
S.~Hochreiter, ``The vanishing gradient problem during learning recurrent
  neural nets and problem solutions,'' \emph{International Journal of
  Uncertainty, Fuzziness and Knowledge-Based Systems}, vol.~06, no.~02, pp.
  107--116, 1998. [Online]. Available:
  \url{https://doi.org/10.1142/S0218488598000094}
\BIBentrySTDinterwordspacing

\bibitem{voskArchitecture}
K.~J. Han, J.~Pan, V.~K.~N. Tadala, T.~Ma, and D.~Povey, ``Multistream cnn for
  robust acoustic modeling,'' 2021.

\bibitem{voskofflinespeechrecognitionapi}
\BIBentryALTinterwordspacing
``Vosk models.'' [Online]. Available: \url{https://alphacephei.com/vosk/models}
\BIBentrySTDinterwordspacing

\bibitem{SnipsNLUPaper}
A.~Coucke, A.~Saade, A.~Ball, T.~Bluche, A.~Caulier, D.~Leroy, C.~Doumouro,
  T.~Gisselbrecht, F.~Caltagirone, T.~Lavril, M.~Primet, and J.~Dureau, ``Snips
  voice platform: an embedded spoken language understanding system for
  private-by-design voice interfaces,'' \emph{CoRR}, vol. abs/1805.10190, 2018.

\bibitem{kuramoto}
F.~A. Rodrigues, T.~K.~D. Peron, P.~Ji, and J.~Kurths, ``The kuramoto model in
  complex networks,'' \emph{Physics Reports}, vol. 610, pp. 1--98, 2016, the
  Kuramoto model in complex networks.

\bibitem{lencioni}
T.~Lencioni, I.~Carpinella, M.~Rabuffetti, A.~Marzegan, and M.~Ferrarin,
  ``Human kinematic, kinetic and emg data during different walking and stair
  ascending and descending tasks,'' \emph{Scientific data}, vol.~6, no.~1, Dec.
  2019.

\bibitem{nuzik}
S.~Nuzik, R.~Lamb, A.~VanSant, and S.~Hirt, ``{Sit-to-Stand Movement Pattern: A
  Kinematic Study},'' \emph{Physical Therapy}, vol.~66, no.~11, pp. 1708--1713,
  11 1986.

\bibitem{SST}
R.~Socher, A.~Perelygin, J.~Wu, J.~Chuang, C.~D. Manning, A.~Y. Ng, and
  C.~Potts, ``Recursive deep models for semantic compositionality over a
  sentiment treebank,'' in \emph{Proceedings of the 2013 conference on
  empirical methods in natural language processing}, 2013, pp. 1631--1642.

\bibitem{GLUE}
\BIBentryALTinterwordspacing
A.~Wang, A.~Singh, J.~Michael, F.~Hill, O.~Levy, and S.~R. Bowman, ``Glue: A
  multi-task benchmark and analysis platform for natural language
  understanding,'' 2018. [Online]. Available:
  \url{https://arxiv.org/abs/1804.07461}
\BIBentrySTDinterwordspacing

\bibitem{nlu-compare}
\BIBentryALTinterwordspacing
X.~Liu, P.~He, W.~Chen, and J.~Gao, ``Multi-task deep neural networks for
  natural language understanding,'' 2019. [Online]. Available:
  \url{https://arxiv.org/abs/1901.11504}
\BIBentrySTDinterwordspacing

\end{thebibliography}

% You can push biographies down or up by placing a \vfill before or after them. The appropriate use of \vfill depends on what kind of text is on the last page and whether or not the columns are being equalized.
%\vfill

% Can be used to pull up biographies so that the bottom of the last one is flush with the other column.

\end{document}